\documentclass{article}
\usepackage{graphicx} 
\usepackage{amsmath} 
\usepackage{amssymb} 
\usepackage{placeins} 
\usepackage{authblk}
\usepackage{xcolor} 
\usepackage{multirow} 
\usepackage{hyperref} 

\title{Sign language recognition from skeletal data using graph and recurrent neural networks}
\author[1,*]{B. Mederos}
\author[2]{J. Mej\'ia}
\author[2]{A. Medina-Reyes}
\author[1]{Y. Espinosa-Almeyda}
\author[2]{J. D. Díaz-Roman}
\author[3]{I. Rodr\'iguez-Mederos}
\author[2]{M. Mej\'ia-Carreon}
\author[1]{F. Gonzalez-Lopez}

\affil[1]{Department of Mathematics, Universidad Aut\'onoma de Ciudad Ju\'arez (UACJ), Ciudad Ju\'arez, Chihuahua, M\'exico. Av. Plutarco Elías Calles No. 1210, Col. Fovissste Chamizal, 32310 Ciudad Juárez, Chihuahua, México}

\affil[2]{Department of Electrical ingeniering, Universidad Aut\'onoma de Ciudad Ju\'arez (UACJ), Ciudad Ju\'arez, Chihuahua, M\'exico. Av. Plutarco Elías Calles No. 1210, Col. Fovissste Chamizal, 32310 Ciudad Juárez, Chihuahua, México}

\affil[3]{Instituto Iberoamericano San Patricio. 
Prof. Aguirre Laredo No. 6142, Col. Partido Iglesias, 32520, Ciudad Juárez, Chihuahua, México }

\affil[*]{Corresponding author: \texttt{\href{mailto:boris.mederos@uacj.mx}{boris.mederos@uacj.mx}}}

\begin{document}
\maketitle
\begin{abstract}
This work presents an approach for recognizing isolated sign language gestures using skeleton-based pose data extracted from video sequences. A Graph-GRU temporal network is proposed to model both spatial and temporal dependencies between frames, enabling accurate classification. The model is trained and evaluated on the AUTSL (Ankara university Turkish sign language) dataset, achieving high accuracy. Experimental results demonstrate the effectiveness of integrating graph-based spatial representations with temporal modeling, providing a scalable framework for sign language recognition. The results of this approach highlight the potential of pose-driven methods for sign language understanding.
\end{abstract}

\section{Introduction}
Sign Language Recognition (SLR) is a growing research area that aims to provide accessible communication for the Deaf community and to support the recognition of its linguistic expressions. Within this field, Isolated Sign Language Recognition (ISLR) focuses on recognizing single, context-free signs, which can be seen as the building blocks for continuous sign language understanding. While ISLR provides an environment to study the visual and temporal characteristics of signs, it also poses significant challenges due to intra-class variations, such as differences in signing speed, hand orientation, and signer-dependent styles. Addressing these challenges is essential for building reliable models that can generalize across diverse signers and recording conditions. Moreover, the recognition of isolated signs provides a basis for Continuous Sign Language Recognition (CSLR), where the boundaries between consecutive signs are not explicitly defined.

ISLR is typically formulated as a classification problem, where each video segment corresponds to a single label representing a sign. Traditional approaches to ISLR \cite{akram2012visual, wang2016islr} focus on handcrafted features or RGB-based video processing. However, these often struggle to generalize due to variations in lighting, background, and signer appearance.  Video-based sign language recognition (SLR) using both RGB and depth data has been widely applied. Reviews such as \cite{adeyanju2021machine} and \cite{al2021deep} summarize key advances and remaining challenges in this area. Other methodologies often rely on 3D convolutional neural networks (3D-CNNs) \cite{Tarres_2023_CVPRW} and, more recently, transformers architectures for spatiotemporal feature modeling \cite{yin2020stmc,Slimane2021}. For instance, \cite{kothadiya2022deepsign} proposes an Indian Sign Language recognition model that combines InceptionResNetV2 for spatial feature extraction with recurrent neural network architectures such as LSTM \cite{hochreiter1997long} and GRU \cite{cho2014gru} networks. Also, \cite{liu2023sign} introduces a Detection Transformer (DETR) framework that couples ResNet152 with multiscale extraction, followed by an encoder-decoder transformer.

Advances in pose estimation \cite{cao2018openpose, kendall2015posenet} have enabled the extraction of skeleton-based representations, in which each sign is modeled as a sequence of joint coordinates. This data representation provides a compact, noise-resistant description of human motion that is suitable for ISLR because it captures both the spatial dependencies between joints and their temporal dynamics. Graph-based neural networks have emerged as a powerful solution to this problem due to their ability to represent human joints as nodes and their anatomical connections as edges. Graph Convolutional Networks (GCNs) \cite{kipf2017semi} can naturally encode the structured topology of the human body \cite{amorim2019spatial}. Although these models were initially developed for action recognition, they have been successfully adapted to sign language tasks, showing superior performance compared to recurrent or convolutional architectures \cite{vazquez2021isolated}. Hybrid approaches that combine GCNs with sequence models have demonstrated that graph-based representations complement temporal modeling, enhancing ISLR  \cite{tunga2021pose}.

Pose-based methods benefit from ignoring irrelevant visual cues, such as background, clothing, or skin tone \cite{Moryossef_2021_CVPR}, and they also reduce model size \cite{gokul2022signpretraining}. For example, \cite{miah2023multistream} constructs skeletal graphs from 27 keypoints across four modalities—joints, joint motion, bones, and bone motion—combining graph convolutions and neural networks for feature extraction and final classification. Likewise, \cite{Laines_2023_CVPRW} proposes converting skeleton keypoints into tree-structured skeleton images and passing them to a DenseNet-121, achieving state-of-the-art performance. Recent improvements combine multi-scale and attention GCNs to capture long-range joint relationships and mitigate over-smoothing in deep graph models \cite{Vazquez-Enriquez_2021_CVPRW, li2022spatial}. These advances illustrate the potential of graph-based methods for  ISLR.

Multimodal strategies have been successfully applied to ISR, boosting recognition performance. the key idea is to integrate RGB, depth, and pose information, fusing predictions across multiple streams \cite{Jiang_2021_CVPR}.  Similarly, the work \cite{chen2022two} proposes a dual-encoder architecture with bidirectional connections and a sign pyramid structure, which processes RGB videos and keypoint sequences jointly, providing promising results.

\section{Methodology}
This section begins by presenting the theoretical foundations of our work on graph and recurrent neural networks. Building upon this foundation, we then introduce our proposed architecture.
\subsection{Graph Neural Networks}
GNNs are designed to process data represented as graphs, where information is associated with nodes and edges. Unlike traditional neural networks, which assume regular grids with images or sequences, GNNs can handle irregular and relational data. The main idea is to iteratively aggregate information from a node's neighbors to update its feature representation. At a particular layer $k$ the feature $\mathbf{h}_v^{(k)}$ of node $v$ is calculated as
\begin{equation}
\mathbf{h}_v^{(k)} = \sigma \Big( \mathbf{W} \cdot \text{AGG}(\{\mathbf{h}_u^{(k-1)} : u \in \mathcal{N}(v)\}) \Big),
\end{equation}
where $\mathcal{N}(v)$ is the set of neighbors, $\mathbf{W}$ is a learnable weight matrix, $\sigma$ is a nonlinear activation, and $\text{AGG}$ represents an aggregation operator. 

The convolution operator, which typically works with grid-like structured data,
is extended to graphs by means of convolutional graph neural networks (GCNs), allowing to capture of relational and topological dependencies between nodes. This type of layer performs a localized spectral convolution by aggregating normalized neighbor features as follows:
\begin{equation}
\mathbf{H}^{(k+1)} = \sigma \Big( \hat{\mathbf{D}}^{-\frac{1}{2}} \hat{\mathbf{A}} \hat{\mathbf{D}}^{-\frac{1}{2}} \mathbf{H}^{(k)} \mathbf{W}^{(l)} \Big),
\end{equation}
where $\hat{\mathbf{A}} = \mathbf{A} + \mathbf{I}$ is the adjacency matrix with self-loops, $\hat{\mathbf{D}}$ is the degree matrix of the graph, $\mathbf{H}^{(k)}$ is a matrix comprising all the node features at layer $k$, and $\mathbf{W}^{(k)}$ is a learnable weight matrix. 

The self-attention mechanism introduced in \cite{vaswani2017attention} has been successfully adapted to graph-structured data. In particular, Graph Attention Networks (GATs) \cite{velickovic2018gat} extend traditional GCNs by incorporating an attention mechanism that allows for updating the nodes' representation as a weighted combination of their neighbors. The weights indicate the contributions (importance) of neighboring nodes to the representation of the node being updated:
\begin{equation}
\mathbf{h}_v^{(k+1)} = \sigma \left( \sum_{u \in \mathcal{N}(v)} \alpha_{vu}^{(k)} \mathbf{W}^{(k)} \mathbf{h}_u^{(k)} \right),
\end{equation}
the attention coefficient between node $v$ and $u\in \mathcal{N}(v)$ is computed as follows:
\begin{equation}
\alpha_{vu}^{(k)}  = \frac{
\exp \big( \text{LeakyReLU}\big( \mathbf{a}^\top [\mathbf{W}^{(k)}\mathbf{h}_v , \mathbf{W}^{(k)}\mathbf{h}_u] \big) \big)
}{
\sum_{k \in \mathcal{N}(v)} 
    \exp \big( \text{LeakyReLU}\big( \mathbf{a}^\top [\mathbf{W}^{(k)}\mathbf{h}_v, \mathbf{W}^{(k)}\mathbf{h}_k] \big) \big)
},
\end{equation}
where $a$ stands for a parameter vector to be learned, $[a,b]$ denotes the concatenation operation between vectors $a$ and $b$, and $\text{LeakyReLU}$ \cite{Xu2015_Empirical} is an activation function that introduces a nonlinearity.
\subsection{Recurrent Neural Networks (RNNs)}
Recurrent Neural Networks (RNNs) are designed to model sequential data, including text, sensor readings, and financial time series. They are widely used in language modeling, speech recognition, and forecasting. Unlike feedforward networks, which assume input independence and cannot retain past information, RNNs incorporate recurrent connections that allow them to capture temporal dependencies and maintain memory across time steps.

At each time step $t$, an RNN maintains a hidden state $\mathbf{h}_t \in \mathbb{R}^h$ that serves as an internal memory summarizing past information. The dynamics of a basic RNN are governed by the following equations:
\begin{equation}
\begin{aligned}
\mathbf{h}_t &= \phi(\mathbf{W}_{hh}\mathbf{h}_{t-1} + \mathbf{W}_{hx}\mathbf{x}_t + \mathbf{b}_h), \\
\mathbf{y}_t &= \psi(\mathbf{W}_{yh}\mathbf{h}_t + \mathbf{b}_y)
\label{rnnequ}
\end{aligned}
\end{equation}
where $\mathbf{x}_t \in \mathbb{R}^d$ is the input, $\mathbf{y}_t \in \mathbb{R}^o$ is the output, $\mathbf{W}_{hh}$, $\mathbf{W}_{hx}$, and $\mathbf{W}_{yh}$ are the recurrent, input, and output weight matrices, and $\mathbf{b}_h$, $\mathbf{b}_y$ are bias vectors. The functions $\phi$ and $\psi$ denote nonlinear activations such as $\tanh$, sigmoid, or ReLU.

For the sake of compactness, equation \eqref{rnnequ} can be rewritten as
\begin{align}
\mathbf{h}_t &= \phi(\mathbf{W}_h[\mathbf{h}_{t-1}, \mathbf{x}_t] + \mathbf{b}_h),\\
\mathbf{y}_t &= \psi(\mathbf{W}_y\mathbf{h}_t + \mathbf{b}_y),
\end{align}
where $[\mathbf{a}, \mathbf{b}]$ denotes the concatenation of $\mathbf{a}$ and $\mathbf{b}$, $\mathbf{W}_h$ combines recurrent and input weights, and $\mathbf{W}_y$ is the output projection matrix.

A standard feedforward network computes, instead,
\begin{align}
\label{eq:ff1}
\mathbf{h}_t &= \phi(\mathbf{W}_{hx}\mathbf{x}_t + \mathbf{b}_h),\\
\mathbf{y}_t &= \psi(\mathbf{W}_{yh}\mathbf{h}_t + \mathbf{b}_y),
\end{align}
which highlights the absence of temporal connections or memory between inputs.

RNNs are typically trained using Backpropagation Through Time (BPTT) \cite{rumelhart1986learning}, 
an extension of standard backpropagation for sequential data. In BPTT, the network is unfolded across time steps, transforming it into an equivalent feedforward network, so that the total loss is expressed as
\begin{equation}
\label{eq:loss}
L(\mathbf{y}, \hat{\mathbf{y}}) = \sum_{t=1}^{T} \ell(\mathbf{y}_t, \hat{\mathbf{y}}_t),
\end{equation}
where $\ell$ is a local loss function, such as mean squared error or cross-entropy,   $\mathbf{y}_t$ and $\hat{\mathbf{y}}_t$ are the expected and predicted network outputs.

However, repeated multiplication of weight matrices across time steps often leads to vanishing or exploding gradients \cite{hochreiter1997long}. Small eigenvalues of $\mathbf{W}_{h}$ cause gradients to decay exponentially, while large ones make the gradients diverge, both affecting long-term learning. To mitigate these problems, gated architectures such as Long Short-Term Memory (LSTM) and Gated Recurrent Units (GRU) introduce control gates to regulate information flow. 

LSTM cells address the vanishing gradient problem by incorporating a memory cell $\mathbf{c}_t$ and three gates: forget $\mathbf{f}_t$, input $\mathbf{i}_t$, and output $\mathbf{o}_t$, that regulate what information is retained, updated, or output. The LSTM equations are:
\begin{eqnarray}
\mathbf{f}_t &=& \sigma(\mathbf{W}_f [\mathbf{h}_{t-1}, \mathbf{x}_t] + \mathbf{b}_f), \\[1mm]
\mathbf{i}_t &=& \sigma(\mathbf{W}_i [\mathbf{h}_{t-1}, \mathbf{x}_t] + \mathbf{b}_i), \\[1mm]
\tilde{\mathbf{c}}_t &=& \tanh(\mathbf{W}_c [\mathbf{h}_{t-1}, \mathbf{x}_t] + \mathbf{b}_c), \\[1mm]
\mathbf{c}_t &=& \mathbf{f}_t \odot \mathbf{c}_{t-1} + \mathbf{i}_t \odot \tilde{\mathbf{c}}_t, \\[1mm]
\mathbf{o}_t &=& \sigma(\mathbf{W}_o [\mathbf{h}_{t-1}, \mathbf{x}_t] + \mathbf{b}_o), \\[1mm]
\mathbf{h}_t &=& \mathbf{o}_t \odot \tanh(\mathbf{c}_t),
\end{eqnarray}
The forget gate $\mathbf{f}_t$ decides which part of the previous cell state $\mathbf{c}_{t-1}$ is retained, the input gate $\mathbf{i}_t$ controls how much new information $\tilde{{\mathbf{c}}}_t$ is added, and the output gate ${\mathbf{o}}_t$ scales the hidden state output. This gating structure allows the network to selectively remember important information over long sequences, enabling effective gradient flow. Intuitively, the ${\mathbf{c}}_t$ acts as a memory, carrying important information forward, while the gates control what to keep, update, or output.

The GRU is a simplified variant of the LSTM that merges the forget and input gates into a single update gate ${\mathbf{z}}_t$, introducing a reset gate ${\mathbf{r}}_t$, which controls the influence of past hidden states on the candidate state. The equations for these gates are stated as follows:
\begin{eqnarray}
\mathbf{z}_t &=& \sigma(\mathbf{W}_z [\mathbf{h}_{t-1}, \mathbf{x}_t] + \mathbf{b}_z), \\[1mm]
\mathbf{r}_t &=& \sigma(\mathbf{W}_r [\mathbf{h}_{t-1}, \mathbf{x}_t] + \mathbf{b}_r), \\[1mm]
\tilde{\mathbf{h}}_t &=& \tanh(\mathbf{W}_h [\mathbf{r}_t \odot \mathbf{h}_{t-1}, \mathbf{x}_t] + \mathbf{b}_h), \\[1mm]
\mathbf{h}_t &=& (1 - \mathbf{z}_t) \odot \mathbf{h}_{t-1} + \mathbf{z}_t \odot \tilde{\mathbf{h}}_t,
\end{eqnarray}
where the update gate $\mathbf{z}_t$ balances how much of the previous hidden state $\mathbf{h}_t$ and the candidate state $\tilde{\mathbf{h}}_t$ contribute to the current state ${\mathbf{h}}_t$. The reset gate $\mathbf{r}_t$ determines how much of the past information should be ignored when computing the candidate $\tilde{\mathbf{h}}_t$. By using these gates, the GRU can capture long-term dependencies with fewer parameters than the LSTM, often leading to faster training with comparable performance.

\subsection{Proposed Architecture}
We define the input as a sequence of graphs over $T$ time steps, where each graph represents the human pose at a specific frame. Let 
$$
\mathcal{G}_t = (\mathbf{V}_t, \mathbf{A}), \quad t = 1, \dots, T
$$
denote the graph at time $t$, with the node feature matrix $\mathbf{V}_t \in \mathbb{R}^{N \times d}$, 
where $N$ is the number of nodes (joint coordinates, velocities, etc.), and each node is represented by a $d$-dimensional feature vector, such that $\mathbf{V}_t \in \mathbb{R}^{N \times d}$, typically $d=2$ or $d=3$ for 2D or 3D coordinates. 
The adjacency matrix $\mathbf{A} \in \{0,1\}^{N \times N}$, representing skeletal connectivity, which remains constant over time. Consequently, the entire sequence can be written as
$$
\mathcal{H}^{(0)} = \{ \mathcal{G}_t \}_{t=1}^T= \{(\mathbf{V}_1, \mathbf{A}), (\mathbf{V}_2, \mathbf{A}) \dots, (\mathbf{V}_T, \mathbf{A})\},
$$
that serves as initial feature space.

At a high level, the proposed architecture processes this input through multiples repeated blocks (stages). Each block combines a GNN, which can be a GCN or GAT, followed by a GRU. The GNN captures the spatial structure of each frame, producing a sequence of graph embeddings, which the GRU then uses to model temporal dependencies across the sequence. Each block incorporates a residual connection similar to those used in ResNet architectures \cite{he2016deep}, allowing the model to preserve and propagate features across blocks. Several blocks are stacked sequentially, after which a final classifier maps the processed features to  200 output classes. 

Specifically, the architecture is organized into $K$ stacked spatio-temporal stages, where each stage operates as a processing block composed of a graph neural network (GNN) followed by a gated recurrent unit (GRU).  That is, the $\ell$ block operates as 
$$
\mathcal{B}^{(\ell)}( \mathcal{H}^{(\ell -1)}) = 
\text{GRU}^{(\ell)}\big(\text{GNN}^{(\ell)}(\mathcal{H}^{(\ell -1)}, \mathbf{A})\big), \ell = 1,\dots,K $$
where $\text{GNN}^{(\ell)}$ processes the spatial structure of each graph in the sequence independently, and $\text{GRU}^{(\ell)}$ captures temporal dependencies across the sequence. 

A residual connection is applied between consecutive blocks, allowing the model to preserve information across stages and providing stability during training; after this, a normalization layer is applied. This is expressed mathematically as 
$$
\mathcal{H}^{(\ell)} =\text{Norm} \left ( (I + \mathcal{B}^{(\ell)})( \mathcal{H}^{(\ell -1)}) \right ) = \text{Norm} \left ( \mathcal{B}^{(\ell)}( \mathcal{H}^{(\ell -1)}) +  \mathcal{H}^{(\ell -1)} \right )
$$
where $I$ denotes the identity. The overall model can be expressed as a composition of residual mappings (stages):
$$
F(\{ \mathcal{G}_t \}_{t=1}^T) 
= 
\left [ \text{Norm} \left ( (I + \mathcal{B}^{(K)}) \right )  \circ \dots \circ \text{Norm} \left ( (I + \mathcal{B}^{(1)}) \right ) \right ](\{ \mathcal{G}_t \}_{t=1}^T),
$$
highlighting the residual structure across the stacked blocks. After processing through all $K$ stages, the output sequence is
$$
\mathcal{H}^{(K)} = \{\mathbf{H}_1^{(K)}, \dots, \mathbf{H}_T^{(L)}\}.
$$
From the sequence $\mathcal{H}^{(K)} \in \mathbb{R}^{B \times T \times N \times H}$, 
where $B$ denotes the batch size, $T$ denotes the number of frames, $N$ denotes the number of joints, and $H$ denotes the feature dimension.  Each temporal slice $\mathbf{H}_t^{(K)} \in \mathbb{R}^{B \times N \times H}$ corresponds to a learned  spatial representation of the pose at time $t$. From this final learned representation, a temporal aggregation attention mechanism is computed, where each spatial embedding $\mathbf{H}_t^{(K)}$ is flattened into vectors $\mathbf{h}_t^{(K)} \in \mathbb{R}^{B \times N \cdot H}$, forming the sequence 
$\mathbf{h}_f^{(K)} = [\mathbf{h}_1^{(K)}, \dots, \mathbf{h}_T^{(K)}]$. The temporal attention mechanism is then applied to emphasize the most informative frames in the sequence. 
For each time step $t$, a scalar score is computed as
$$
e_t = \mathbf{w}_t^{\top} \mathbf{h}_t^{(K)} + b_t,
$$
using a linear layer for learning the weights $\mathbf{w}_a$; these scores are normalized across all frames using a softmax function to produce attention weights
$$
\alpha_t = \frac{\exp(e_t)}{\sum_{k=1}^{T} \exp(e_k)}.
$$

The final sequence representation is obtained as a weighted sum of the temporal embeddings:
$$
\mathbf{h}_{\text{reduced}} = \sum_{t=1}^{T} \alpha_t \, \mathbf{h}^{(K)}_t,
$$
yielding a compact feature vector that summarizes the temporal dynamics of the input sequence. 
This aggregated representation is subsequently passed through a multi-layer classifier, consisting of two fully connected layers interleaved with dropouts, producing $\mathbf{h}_{\text{final}}$, which is fed into the last classifier layer
$$
\hat{\mathbf{y}} = \text{softmax}\big(\mathbf{W}_c \mathbf{h}_{\text{final}} + \mathbf{b}_c\big).
$$

\section{Results}
The proposed model was trained with the AUTSL dataset \cite{ozdemir2020autsl}, composed of turkish sign videos. The AUTSL dataset was considered for its size, signer diversity, and class balance. In total, the dataset contains 226 distinct sign classes performed by 43 different signers, providing significant variability in signer gesture execution. 
The dataset was divided into three subsets: training, validation, and test, comprising
28143, 4419, and 3743 RGB videos, respectively, where each video has an average duration of 2.05 seconds. A key advantage of AUTSL is the number of available samples for each class, enabling training without severe class imbalance.

The videos  were fed to the PoseNet model \cite{toshev2014deeppose}  to extract the 2D skeletal keypoints of the signer for each video frame. These keypoints represent the spatial coordinates of the upper-body joints $(x,y)$, along with a confidence score $c$ for each detected point, which was removed from the training process due to its irrelevance as determined by experimentation.  

The proposed model was trained using the new keypoint representation. The corresponding architecture consisted of 16 stages, where each stage is composed of a GAT followed by a GRU layer, and the number of heads in the GAT is 8. The training was carried out over 100 epochs using a batch size of 64. The AdanW optimizer \cite{loshchilov2019adamw} was employed with an initial learning rate of $10^{-3}$ and a weight decay of $10^{-5}$. 

\begin{figure}[h]
    \centering
    \includegraphics[width=0.7\textwidth]{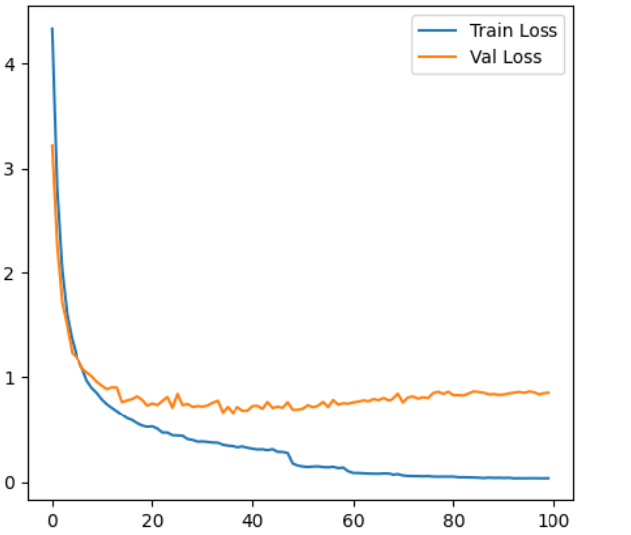}
    \caption{Evolution of the train and validation loss during trainning}
    \label{fig1:imagen}
\end{figure}

\begin{figure}[h]
    \centering \includegraphics[width=0.7\textwidth]{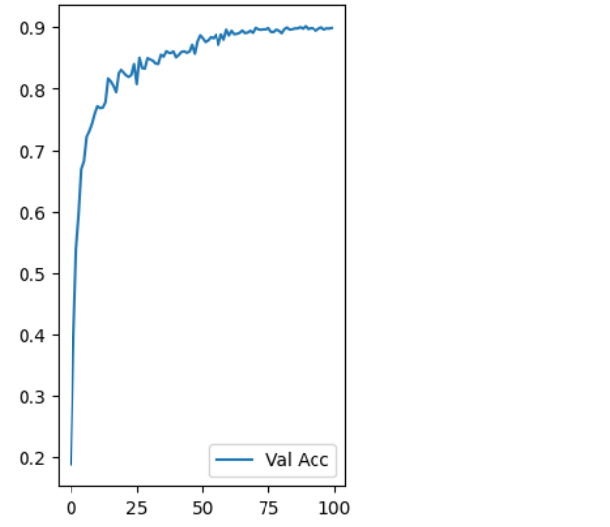}
    \caption{Evolution of the train and validation loss during trainning}
    \label{fig2:accuracy}
\end{figure}

The decay of the loss in both the training and validation datasets is shown in figure \ref{fig1:imagen}, and the validation accuracy is depicted in \ref{fig2:accuracy}, reaching an accuracy of  90.04\%

\FloatBarrier

\FloatBarrier

The proposed architecture outperforms both the video-based ResNet2D+1 \cite{Tran_2018_CVPR} and the skeleton-based Conv1D+GRU  ensemble model \cite{medina2025sign} with respect to the accuracy metric, as shown in Table~\ref{tab:comparison_results}. Moreover, it also exhibits a clear advantage in computational efficiency. While ResNet2D+1 requires over 43 hours of training and 197.6s per inference, our model achieves faster performance comparable to that of the Conv1D+GRU ensemble model. This is due to its compact pose-based input representation, which drastically reduces computational and memory demands.

\begin{table}[h]
\centering
\label{tab:comparison_results}
\begin{tabular}{c|c|c|c|c}
\hline
\multirow{2}{*}{\textbf{Model}} & 
\multirow{2}{*}{\textbf{Input}}  & \multirow{2}{*}{\textbf{Acc}} & \textbf{Training} & \textbf{Inference} \\
 &   &  & \textbf{time} (h)& \textbf{time} (s)\\
\hline
RESNET2D+1 & Images & 88.90\% & 43.8 & 197.6 \\
Conv1D+GRU ensemble & Skeleton data   & 88.32\% & 1.0 & 0.95 \\
Proposed Architecture &  Skelenton data  & 90.04\% & 1.01 & 0.95 \\
\hline
\end{tabular}
\caption{Comparison among the proposed architecture, RESNET2D+1, and Conv1D+GRU ensemble in terms of accuracy (\%), training time (h), and inference time (s).}
\end{table}
\FloatBarrier

The Table \ref{tab1} shows the top 20 classes by precision, where several classes achieve a perfect precision of 1.0, indicating that whenever the model predicts these classes, it is always correct. However, some classes, such as class 6, have a much lower recall ($\approx0.23$), meaning that the model misses many true instances despite being accurate when it predicts them. Other classes, such as 32, 14, or 27, show both high precision and recall, reflecting that the model is able to detect and correctly classify these signs. In contrast, Table \ref{tab2} shows the 20 classes with the lowest precision ($\approx 0.51–0.72$), revealing that the model often confuses these classes with others. Some of these low-precision classes have high recall, meaning that most true instances are detected; however, many predictions are assigned incorrectly.  Overall, these Tables  highlight the model’s strengths in recognizing certain classes and its weaknesses in handling the more confusing or difficult classes. 

\begin{table}[h!]
\centering
\caption{Precision, Recall and F1-score for top classes}
\label{tab:metrics}
\begin{tabular}{c|c|c|c|c|c|c|c|}
\hline
Class & Precision & Recall & F1-score & Class & Precision & Recall & F1-score \\
15  &  1.000000  &  1.00 & 1.000000  &106 &  1.000000  &  1.00 & 1.000000 \\
14  &  1.000000  &  1.00 & 1.000000  &120 &  1.000000  &  1.00 & 1.000000  \\
27  &  1.000000  &  1.00 & 1.000000  &160 &  1.000000  &  1.00 & 1.000000  \\
32  &  1.000000  &  1.00 & 1.000000  &114 &  1.000000  &  1.00 & 1.000000  \\
11  &  1.000000  &  1.00 & 1.000000  &118 &  1.000000  &  1.00 & 1.000000  \\
25  &  1.000000  &  1.00 & 1.000000  &119 &  1.000000  &  1.00 & 1.000000  \\
88  &  1.000000  &  1.00 & 1.000000  &116 &  1.000000  &  1.00 & 1.000000  \\
69  &  1.000000  &  1.00 & 1.000000  &39  &  0.952381  &  1.00 & 0.975610  \\
72  & 1.000000   & 1.00  & 1.000000  &2   &  0.952381  &  1.00 & 0.975610  \\
147 & 1.000000   & 1.00  & 1.000000  &217 &  0.952381  &  1.00 & 0.975610  \\
139 & 1.000000   & 1.00  & 1.000000  &38  &  0.952381  &  1.00 & 0.975610  \\
214 & 1.000000   & 1.00  & 1.000000  &141 &  0.952381  &  1.00 & 0.975610  \\
222 & 1.000000   & 1.00  & 1.000000  &152 &  0.952381  &  1.00 & 0.975610  \\
219 & 1.000000   & 1.00  & 1.000000  &223 &  0.952381  &  1.00 & 0.975610  \\
210 & 1.000000   & 1.00  & 1.000000  &170 &  0.950000  &  1.00 & 0.974359  \\
193 & 1.000000   & 1.00  & 1.000000  &113 &  1.000000  &  0.95 & 0.974359  \\
203 &  1.000000  &  1.00 & 1.000000  &133 &  1.000000  &  0.95 & 0.974359  \\
196 &  1.000000  &  1.00 & 1.000000  &102 &  0.950000  &  1.00 & 0.974359  \\
168 &  1.000000  &  1.00 & 1.000000  &110 &  1.000000  &  0.95 & 0.974359  \\
167 &  1.000000  &  1.00 & 1.000000  &70  &  1.000000  &  0.95 & 0.974359  \\
174 &  1.000000  &  1.00 & 1.000000  &41  &  1.000000  &  0.95 & 0.974359  \\
179 &  1.000000  &  1.00 & 1.000000  &60  &  1.000000  &  0.95 & 0.974359  \\
180 &  1.000000  &  1.00 & 1.000000  &31  &  1.000000  &  0.95 & 0.974359  \\
155 &  1.000000  &  1.00 & 1.000000  &83  &  1.000000  &  0.95 & 0.974359  \\
151 &  1.000000  &  1.00 & 1.000000  &71  &  1.000000  &  0.95 & 0.974359  \\
\hline
\end{tabular}
\end{table}

\begin{table}[h!]
\centering
\caption{Precision, Recall and F1-score for top 20 classes}
\label{tab2}
\begin{tabular}{c|c|c|c|c|c|c|c}
\hline
Clase & Precision & Recall & F1-score & Clase & Precision & Recall & F1-score\\
\hline
144 &  0.750000 & 0.900000 & 0.818182 & 166 & 0.714286 & 0.750000 & 0.731707  \\
105 &  0.689655 & 1.000000 & 0.816327 & 79  & 0.566667 & 1.000000 & 0.723404 \\
186 &  0.882353 & 0.750000 & 0.810811 & 36  & 0.764706 & 0.684211 & 0.722222 \\
171 &  0.703704 & 0.950000 & 0.808511 & 86  & 0.736842 & 0.700000 & 0.717949 \\
148 &  0.720000 & 0.900000 & 0.800000 & 131 & 0.833333 & 0.625000 & 0.714286 \\
103 &  0.800000 & 0.800000 & 0.800000 & 135 & 0.764706 & 0.650000 & 0.702703 \\
0   &  0.800000 & 0.800000 & 0.800000 & 93  & 0.652174 & 0.750000 & 0.697674 \\
68  &  1.000000 & 0.666667 & 0.800000 & 154 & 0.800000 & 0.600000 & 0.685714 \\
10  &  0.789474 & 0.789474 & 0.789474 & 67  & 0.722222 & 0.650000 & 0.684211 \\
128 &  1.000000 & 0.650000 & 0.787879 & 178 & 0.625000 & 0.750000 & 0.681818 \\
143 &  0.645161 & 1.000000 & 0.784314 & 127 & 0.625000 & 0.750000 & 0.681818 \\
184 &  0.692308 & 0.900000 & 0.782609 & 9   & 0.650000 & 0.684211 & 0.666667 \\
165 &  0.875000 & 0.700000 & 0.777778 & 183 & 0.548387 & 0.850000 & 0.666667 \\
8   &  0.875000 & 0.700000 & 0.777778 & 52  & 0.515152 & 0.944444 & 0.666667 \\
51  &  0.923077 & 0.666667 & 0.774194 & 65  & 0.705882 & 0.600000 & 0.648649 \\
169 &  0.625000 & 1.000000 & 0.769231 & 22  & 0.705882 & 0.600000 & 0.648649 \\
17  &  0.812500 & 0.722222 & 0.764706 & 16  & 0.909091 & 0.500000 & 0.645161 \\
47  &  0.928571 & 0.650000 & 0.764706 & 95  & 0.857143 & 0.500000 & 0.631579 \\
126 &  0.750000 & 0.750000 & 0.750000 & 122 & 0.666667 & 0.600000 & 0.631579 \\
140 &  1.000000 & 0.600000 & 0.750000 & 111 & 1.000000 & 0.450000 & 0.620690 \\
172 &  0.750000 & 0.750000 & 0.750000 & 221 & 0.647059 & 0.550000 & 0.594595 \\
90  &  0.750000 & 0.750000 & 0.750000 & 97  & 0.647059 & 0.550000 & 0.594595 \\
149 &  0.812500 & 0.684211 & 0.742857 & 96  & 0.727273 & 0.470588 & 0.571429 \\
91  &  0.777778 & 0.700000 & 0.736842 & 98  & 0.800000 & 0.421053 & 0.551724 \\
187 &  1.000000 & 0.578947 & 0.733333 & 6   & 1.000000 & 0.230769 & 0.375000 \\
\hline
\end{tabular}
\end{table}

\section{Conclusion}
This work proposes a hybrid model combining Graph Neural Networks and Gated Recurrent Units for isolated sign language recognition using skeletal data. The approach models each sign as a temporal sequence of graphs, effectively capturing both spatial dependencies between joints and temporal motion dynamics. Additionally, residual connections and temporal attention enhance feature preservation and focus on informative frames, leading to stable convergence and competitive accuracy. The results confirm that integrating graph and recurrent modeling provides a compact yet powerful representation compared to traditional methods. Future extensions will address continuous sign recognition and explore multimodal and transformer-based architectures to improve robustness and generalization.

\subsection*{Acknowledgement}
Y. Espinosa-Almeyda acknowledges SECIHTI for the postdoctoral scholarship ``Estancias Postdoctorales por M\'exico para la Formaci\'on y Consolidaci\'on de Investigadores por M\'exico" at IIT, UACJ, 2024-2026. All authors thank the SECIHTI for the financial supports.

\section{Bibliography}
\bibliographystyle{ieeetr} 
\bibliography{bibliography}
\end{document}